# Synthetic Survival Data Generation for Heart Failure Prognosis Using Deep Generative Models

Chanon Puttanawarut[1,2]*, Natcha Fongsrisin[3], Porntep Amornritvanich[4,2], Panu Looareesuwan[2]*, Cholatid Ratanatharathorn[2]

1. Chakri Naruebodindra Medical Institute, Faculty of Medicine Ramathibodi Hospital, Mahidol University, Samut Prakan, Thailand
2. Department of Clinical Epidemiology and Biostatistics, Faculty of Medicine Ramathibodi Hospital, Mahidol University, Bangkok, Thailand
3. Department of Systems Design Engineering, University of Waterloo, Waterloo, Canada
4. Cardiovascular Unit, Department of Internal Medicine, Police General Hospital, Bangkok Thailand

**Abstract**

**Background**: Heart failure (HF) research is constrained by limited access to large, shareable datasets due to privacy regulations and institutional barriers. Synthetic data generation offers a promising solution to overcome these challenges while preserving patient confidentiality.
**Methods**: We generated synthetic HF datasets from institutional data comprising 12,552 unique patients using five deep learning models: tabular variational autoencoder (TVAE), normalizing flow, ADSGAN, SurvivalGAN, and tabular denoising diffusion probabilistic models (TabDDPM). We comprehensively evaluated synthetic data utility through statistical similarity metrics, survival prediction using machine learning and privacy assessments.
**Results**: SurvivalGAN and TabDDPM demonstrated high fidelity to the original dataset, exhibiting similar variable distributions and survival curves after applying histogram equalization. SurvivalGAN (C-indices: 0.71-0.76) and TVAE (C-indices: 0.73-0.76) achieved the strongest performance in survival prediction evaluation, closely matched real data performance (C-indices: 0.73-0.76). Privacy evaluation confirmed protection against re-identification attacks.
**Conclusions**: Deep learning-based synthetic data generation can produce high-fidelity, privacy-preserving HF datasets suitable for research applications. This publicly available synthetic dataset addresses critical data sharing barriers and provides a valuable resource for advancing HF research and predictive modeling.

**\*Corresponding Authors:**

Cholatid Ratanatharathorn: rcholatid@gmail.com

Chanon Puttanawarut: chanonp@protonmail.com

## 1. Introduction

Heart failure (HF) represents one of the most significant global health challenges, affecting over 64 million people worldwide with prevalence projected to increase by 46% by 2030 [1,2]. Despite advances in cardiovascular therapeutics, HF mortality continues to rise, with recent data showing 425,147 annual deaths in the United States alone [3]. These trends highlight the need for research that can comprehensively characterize HF across diverse patient populations, supported by large, high-quality datasets.

Artificial intelligence (AI) has shown potential in addressing these challenges and is increasingly recognized for its value in healthcare [4], yet their successful implementation depends on access to large, high-quality, and representative datasets. Traditional clinical trials, while providing gold-standard evidence, are resource-intensive, demanding substantial time, personnel, and financial investment [5,6]. Consequently, there is growing interest in leveraging real-world clinical data to develop and validate predictive models that can inform clinical decision-making and identify patients at highest risk.

Validating results through the use of external datasets from different clinical settings remains a significant challenge, primarily due to the sensitive nature of patient data. Barriers to data sharing add further obstacles to HF research. Complex legal frameworks across multiple regulations create obstacles to collaborative research, while patient privacy concerns and institutional data custodianship policies default to restrictive sharing practices [7,8]. Traditional anonymization techniques such as deidentification exhibit vulnerabilities, including susceptibility to linkage attacks, where anonymized records are cross-referenced with external datasets and statistical inference attacks that can re-identify individuals or expose sensitive attributes [9,10]. Furthermore, these approaches consistently demonstrate trade-offs between privacy protection and data utility, with anonymization processes resulting in substantial information loss that can compromise the validity of downstream analyses [11]. The emergence of federated learning approaches offers partial solutions, but implementation remains limited by technical infrastructure requirements and regulatory uncertainty [7,12].

Synthetic data generation through generative artificial intelligence offers a solution to these challenges. Recent advances in deep learning have demonstrated the capability to generate clinically realistic tabular patient data while preserving privacy [13–17]. Previous studies also show successful implementations of synthetic tabular data in medicine, such as heart failure [18], acute myeloid leukemia [14], small cell lung cancer [19], breast cancer [19], and diabetes [19]. While generative adversarial networks (GANs) have been the most widely adopted architecture in medicine, the potential of other advanced models, such as variational autoencoders, normalizing flows, and diffusion models, remains less explored for complex clinical datasets [16,20,21].

To address these problems, the aim of this study is to produce a high-fidelity, publicly shareable synthetic dataset for heart failure research. We apply deep learning models to create synthetic tabular HF data that can be made publicly available. Using comprehensive institutional data from 12,552 unique patients, we explore multiple deep learning approaches for synthetic data generation. Our objective is to create a high-fidelity, publicly shareable synthetic dataset that can accelerate HF research while evaluating both the analytical utility and privacy preservation characteristics of the generated data.

## 2. Material and Methods

The overall methodology for generating and evaluating synthetic datasets is summarized in Figure 1. The process involved preprocessing the raw data, training five distinct generative models, and subsequently assessing the resulting synthetic data for utility and privacy. The following sections provide a detailed account of each step.

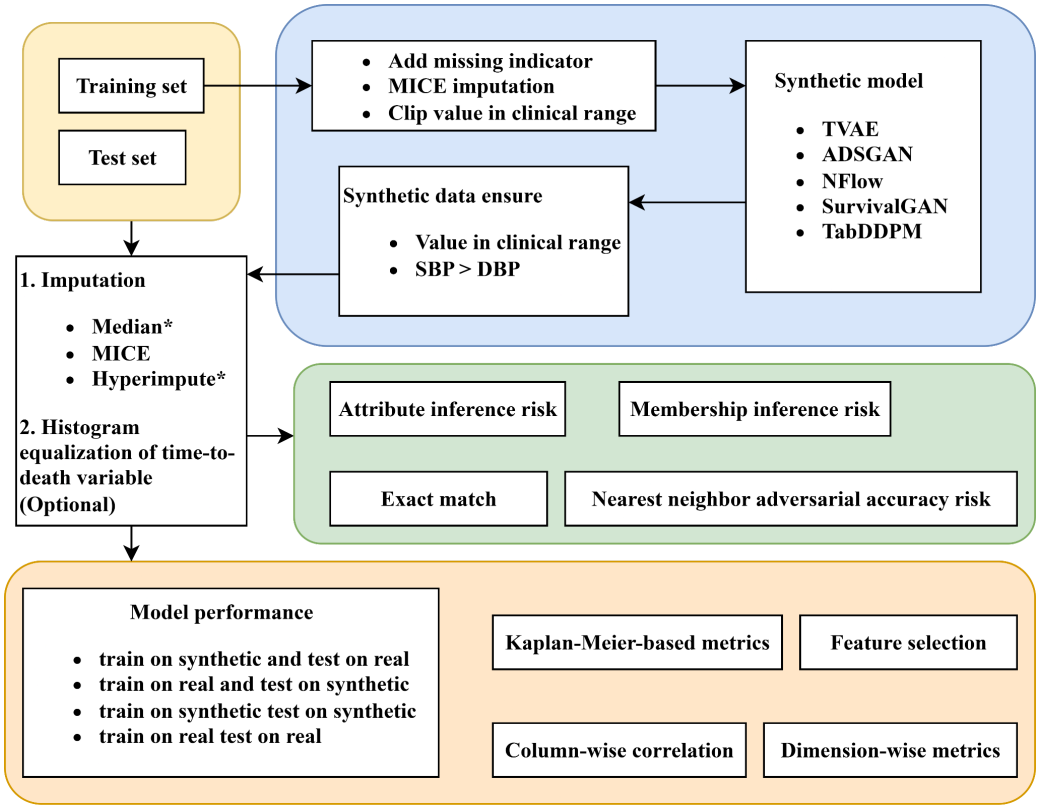

Figure 1: Overall schematic of the study methodology. The flowchart illustrates the pipeline for synthetic data generation, imputation, and the framework assessing data utility and privacy risks.

## 2.1 Dataset

Our study uses a dataset of a real-world retrospective cohort of 12,552 patients with HF, collected at Ramathibodi Hospital between January 2010 and June 2025. The study population was identified through hospital records and electronic health databases, including ICD-10 [22] codes, medication records, and echocardiographic reports. Supplementary information such as medication prescriptions, comorbidities, laboratory results, and procedures was extracted from the hospital information system. Ethical approval for this study was obtained from the Institutional Review Board of the Faculty of Medicine, Ramathibodi Hospital prior to initiation of the research (MURA2025/121). For more information about the patient identification process and cohort construction, please refer to the Supplementary D.

## 2.2 Data Preprocessing

To protect patient privacy, the dataset was de-identified in accordance with the health insurance portability and accountability act (HIPAA) Privacy Rule. We remove all 18 types of data according to HIPAA [23]. Then the original dataset was partitioned into a training set and a test set. Initial preprocessing was applied to the training data before its use in the synthetic models. To handle missing values, we first added a missing value indicator column (1=present, 0=missing) for each feature containing nulls (see Supplementary Table 1 for details on missingness). Subsequently, multivariate imputation by chained equations (MICE) was employed to impute the missing data points on the training set. Finally, to handle outliers and ensure values remained within plausible clinical ranges, we clipped the data to predetermined minimum and maximum values by cardiologist (Supplementary table 2).

## 2.3 Synthetic Data Generation

Five synthetic data generation models were trained on the preprocessed real data to produce five distinct synthetic datasets: tabular variational autoencoder (TVAE) [24], normalizing flow (NFlow) [25], ADSGAN (anonymization through data synthesis using generative adversarial networks) [26], SurvivalGAN (a GAN architecture tailored for survival data) [27], and tabular denoising diffusion probabilistic models (TabDDPM). These models were implemented in Python using the Synthcity library [28] with their default hyperparameters. For additional information about the development of synthetic data, I refer to Supplementary section B.

Following the generation process, a filtering step was applied to ensure the clinical and logical integrity of the data. This step removed any synthetic records that violated predefined constraints, such as a systolic blood pressure (SBP) value being less than or equal to the diastolic blood pressure (DBP) or values falling outside of clinically plausible ranges (Supplementary table 2).

## 2.4 Evaluation

We evaluated the generated synthetic data using a comprehensive framework that assessed three critical dimensions including statistical fidelity, utility for downstream machine learning tasks, and privacy preservation. Statistical fidelity measures how well synthetic data preserves distributional and structural properties of the original data. To deal with missing values, all datasets underwent preprocessing with MICE imputation, generating a single complete dataset per original dataset using a maximum of 100 iterations for convergence. This standardized preprocessing step was applied uniformly to both real and synthetic datasets before conducting any evaluations. To ensure Kaplan-Meier curve similarity, we also conducted an ablation experiment comparing synthetic datasets with and without histogram equalization applied to the time-to-death variable ("Days") to match the distribution observed in the real training data.

**Utility Evaluation**

### Dimension-wise Distribution

We assessed how well synthetic datasets replicated the marginal distributions of original data using feature-specific metrics. For categorical features, we measured similarity using 1 minus the mean absolute frequency difference between real and synthetic data. For continuous features, we applied the two-sample Kolmogorov-Smirnov test, reporting similarity as 1 minus the KS statistic (which measures the maximum difference between cumulative distribution functions). Both normalized scores range from 0 to 1, where higher values indicate better distributional fidelity.

### Column-wise Correlation

This metric quantifies how well the synthetic dataset preserves the inter-feature correlation structure of the real dataset. For continuous variable pairs, we computed Pearson correlation coefficients for both real and synthetic data, then derived a similarity score using the formula: score = $1 - |S - R|/2$, where S and R represent synthetic and real correlations respectively. For categorical variable pairs, we compared joint probability distributions by constructing normalized contingency tables and computing similarity using distance score calculated by: $1 - \frac{1}{2}\sum|S_{xy} - R_{xy}|$, where x and y represent all possible category combinations. Both metrics yield scores from 0 (completely different) to 1 (identical), capturing whether synthetic data maintains the same relational patterns as the original data.

### Feature selection

To assess whether synthetic datasets preserved feature importance patterns from real data, we employed Cox proportional hazards (CoxPH) models on the entire dataset without preprocessing. For each dataset (real and synthetic), we fitted univariate CoxPH models for all available features and considered a feature preserved if it showed statistical significance ($p < 0.05$) with the same sign as in the real data.

### Machine Learning Performance

The most critical utility test involved measuring predictive model performance when trained on synthetic data. We employed three evaluation paradigms: train-on-synthetic/test-on-real (TSTR), train-on-real/test-on-synthetic (TRTS), and train-on-synthetic/test-on-synthetic (TSTS). Model performance was quantified using two complementary metrics. The concordance index (C-index) [29] assessed discriminative ability, the model's capacity to correctly rank subjects by predicted survival risk. The integrated Brier score (IBS) measured mean squared error between predicted survival probabilities and actual outcomes across the 10th to 90th percentiles of follow-up time, providing a comprehensive assessment of model calibration and accuracy (lower values indicate better performance) [30,31].

We evaluated four distinct model architectures to ensure robust assessment across different machine learning paradigms: CoxPH, Random Survival Forest (RSF) [32], DeepSurv [33], and DeepHit [34]. CoxPH served as the traditional semi-parametric baseline, modeling the hazard as a product of a baseline hazard and a linear combination of covariates without assuming a specific baseline distribution. RSF extends random forests to survival analysis by constructing multiple survival trees and aggregating their predictions. DeepSurv adapts the CoxPH model into a neural network, using feedforward layers to capture non-linear relationships while preserving the partial likelihood framework. DeepHit, in contrast, is a fully parametric deep learning model that directly estimates discrete survival distributions without relying on proportional hazards assumptions.

To ensure robustness of machine learning performance evaluation, we tested additional imputation strategies beyond MICE, including simple median imputation and the HyperImpute method [35], with default hyperparameters using HyperImpute library, allowing comparison of synthetic data utility across different data preprocessing approaches. For additional information about the development of machine learning, I refer to Supplementary section C.

### Kaplan-Meier-based Metrics

To specifically assess the utility for the time-to-event and censoring distributions, we employed three KM based metrics introduced by Norcliffe et. al. [27]. for evaluating synthetic survival data. These metrics included optimism, which measures the difference in expected lifetime between synthetic and real data by comparing the areas under their respective KM curves. KM divergence quantifies the mean absolute difference between survival curves. Short-sightedness evaluates whether the synthetic data captures the full temporal range of the original dataset by comparing the end times of the survival curves. For all three metrics, values closer to zero indicate superior correspondence between the survival curves derived from synthetic and real data.

**Privacy Evaluation**

**Exact Match**

This analysis was conducted to test for exact matches between real and synthetic records, aiming to detect any direct replication from the original dataset. A synthetic record was considered a "match" if it closely resembled a real one based on a composite rule: categorical values had to match exactly, while numerical values were considered matching if they differed by no more than 5% from the real value. The goal is to observe zero matches, which would indicate that no real records were directly copied into the synthetic dataset.

**Membership Inference Attack (MIA)**

We evaluated privacy leakage by implementing a membership inference attack that determines whether specific records were included in the training data used to generate synthetic datasets. Our approach uses a distance-based classifier that measures the Euclidean distance between each record and its nearest neighbor in the synthetic dataset [13,36]. After normalizing all datasets using a shared min and max value fitted on the combined training and synthetic data, we fit a k-nearest neighbors model (k=1) on the synthetic data and compute distances from both training set (member) and test set (non-member) to their closest synthetic counterparts. Records with distances below a threshold, set at the 50th percentile of the combined distance distribution, are classified as members. We employed 4-fold cross-validation to balance the dataset sizes, ensuring the number of training sets (members) match the number of test sets (non-members) for fair comparison. The reported MIA accuracy represents the average performance across all four folds, where values approaching 0.5 indicate strong privacy protection (equivalent to random guessing).

**Attribute Inference Attack (AIA)**

This test evaluates the risk of an attacker inferring sensitive attributes from real data using synthetic datasets as auxiliary information. We defined quasi-identifier columns (height, body weight, age, gender) and treated all remaining features as sensitive attributes that an attacker might attempt to infer. Using 5-fold cross-validation on each synthetic dataset, for each fold, we trained machine learning models on the synthetic data to predict each sensitive attribute using the remaining features as predictors. For categorical attributes, we use logistic regression (AIA Linear) and XGBoost (AIA XGB) models, measuring prediction accuracy with label-encoded targets. For continuous attributes, we use linear regression (AIA Linear) and XGBoost (AIA XGB) models, evaluating proximity using relative differences of 5% to determine if predictions were sufficiently close to true values. The trained models were then evaluated on real training data folds to measure their predictive performance, simulating an attacker's ability to infer sensitive information about real individuals when given partial knowledge and access to synthetic data. We computed the average accuracy score across all fold and all sensitive features to obtain a single privacy risk metric per dataset. For comparison, we also evaluated the same models on held-out real test data to establish a baseline. Scores closer to 0.5 indicate lower privacy risk.

**Nearest Neighbor Adversarial Accuracy (NNAA)**

The NNAA metric quantifies the statistical distinguishability between datasets by comparing nearest-neighbor distances within and across dataset boundaries using Euclidean distance on normalized features [13,37]. All variables are normalized using min-max scaling prior to distance computation. The adversarial score between any two datasets p and q is computed as:

$$\text{NNAA}(p,q) = 0.5 \times \left( \frac{1}{|p|} \sum_{x \in p} \mathbf{1}[d(x, \text{NN}_q(x)) > d(x, \text{NN}_p(x))] + \frac{1}{|q|} \sum_{y \in q} \mathbf{1}[d(y, \text{NN}_p(y)) > d(y, \text{NN}_q(y))] \right)$$

where $d(\cdot, \cdot)$ represents the Euclidean distance, $\text{NN}_p(x)$ denotes the nearest neighbor of point $x$ within dataset $p$, and $|p|$ denotes the size of dataset $p$. When finding nearest neighbors within the same dataset (i.e., $\text{NN}_p(x_i)$ where $x_i \in p$), the algorithm excludes the query point itself to prevent zero distances from self-matching. The nearest neighbor is determined by finding the point with minimum Euclidean distance in the normalized feature space.

The evaluation involves two primary comparisons: NNAA(T,S) measures distinguishability between the training set T and synthetic set S, while NNAA(E,S) assesses how well the synthetic data represents the holdout test set E. The privacy loss is calculated as: Privacy Loss = NNAA(E,S) - NNAA(T,S), where a positive value suggests potential overfitting of the synthetic data to the training set. When datasets differ in size, the metric employs resampling to ensure balanced comparisons, computing the average across 30 iterations. An ideal Privacy Loss of 0.0 signifies that the model's ability to create indistinguishable data is consistent for both the data it was trained on and new, unseen data.

## 3. Results

We generated 12,552 synthetic patients for each model to match the size of the real dataset. The summary statistics of real dataset and synthetic dataset prior to imputation of missing data are presented in Supplementary Table 1. Overall, the synthetic data produced by all models closely resembled the real data, as reflected by the alignment of their distributions along the diagonal line in Figure 2. For continuous variables, TabDDPM most accurately replicated the original distributions, followed by ADSGAN and SurvivalGAN (Supplementary Figures 2-6). In contrast, TVAE produced characteristically spiky and non-smooth distributions, while NFlow exhibited the most substantial divergence in the distribution of the 'Days' variable, an important variable for survival analysis. The percentage of missing data for continuous variables showed good agreement between synthetic and real data, as indicated by alignment with the diagonal line (Supplementary Figure 1).

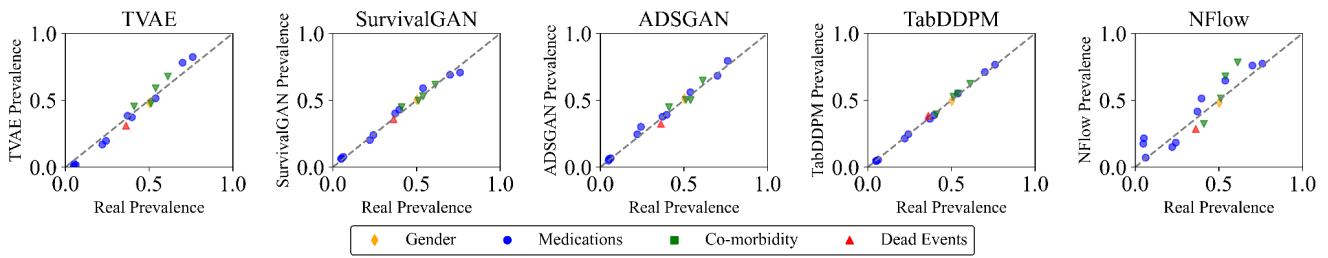

Figure 2: Comparison of the prevalence of categorical variables between the real data (x-axis) and synthetic data (y-axis). Each point represents a feature, categorized by type. The dashed line indicates a perfect correspondence.

For most variables, the means of values between real and synthetic datasets were similar, although statistical testing revealed significant differences (Supplementary Table 1). This finding must be interpreted with caution. Given the large sample size of 12,552, even small differences can appear statistically significant. TabDDPM produced the highest number of variables that were not statistically different from the real data (22). This conclusion was based on appropriate statistical tests with a significance level of $p<0.05$. The tests included Welch's t-test or Mann-Whitney U test for continuous variables, Fisher's exact test for binary categorical variables, and the chi-square test for non-binary categorical variables.

All models achieved over 80% accuracy in dimension-wise metrics and column-wise correlations, with TabDDPM performing best (dimension-wise: 0.979, column-wise: 0.949) (Figure 3A). TabDDPM's KM curves most closely matched real data (Figure 3B) with low optimism and KM divergence (<0.03) and near-zero short-sightedness, while TVAE, NFlow, ADSGAN, and SurvivalGAN showed greater divergence (Figure 3C). For prognostic feature preservation, SurvivalGAN achieved highest recall (0.87) in identifying significant features with consistent risk/protective effects, while maintaining precision of 0.83, and TVAE achieved highest precision (0.947) with recall of 0.78 for detecting significant features with the same directional association (risk or protective) as the real dataset (Figure 3D). Most models maintained recall/precision ≥0.7 for preserving these directional relationships except NFlow and ADSGAN, whose recall is 0.61 and 0.65, respectively.

The heatmap analysis of log hazard ratios for mortality risk (Supplementary Figure 14) revealed that all synthetic models produced visually similar patterns to the real dataset. However, three models (NFlow, ADSGAN, and TabDDPM) showed a few cases of directionally opposite significant values compared to the real dataset (CKD in NFlow and ADSGAN, ACEI in NFlow, and potassium in TabDDPM). Across synthetic models, TVAE shows the most clinically coherent set of significant signals. It has significant protection for beta-blockers, SGLT2i, ARBs, MRAs, statins, and systolic blood pressure, and significant harm for atrial fibrillation, diabetes, and ivabradine. SurvivalGAN looks closest to the real dataset. It keeps several significant protective markers such as beta-blockers, SGLT2 inhibitors, hemoglobin, and potassium, but it also shows significant harm for ARNI and ivabradine. Consider only in heatmap, prefer models where the main clinical effects are significant in the expected direction. TVAE and SurvivalGAN meet this need best.

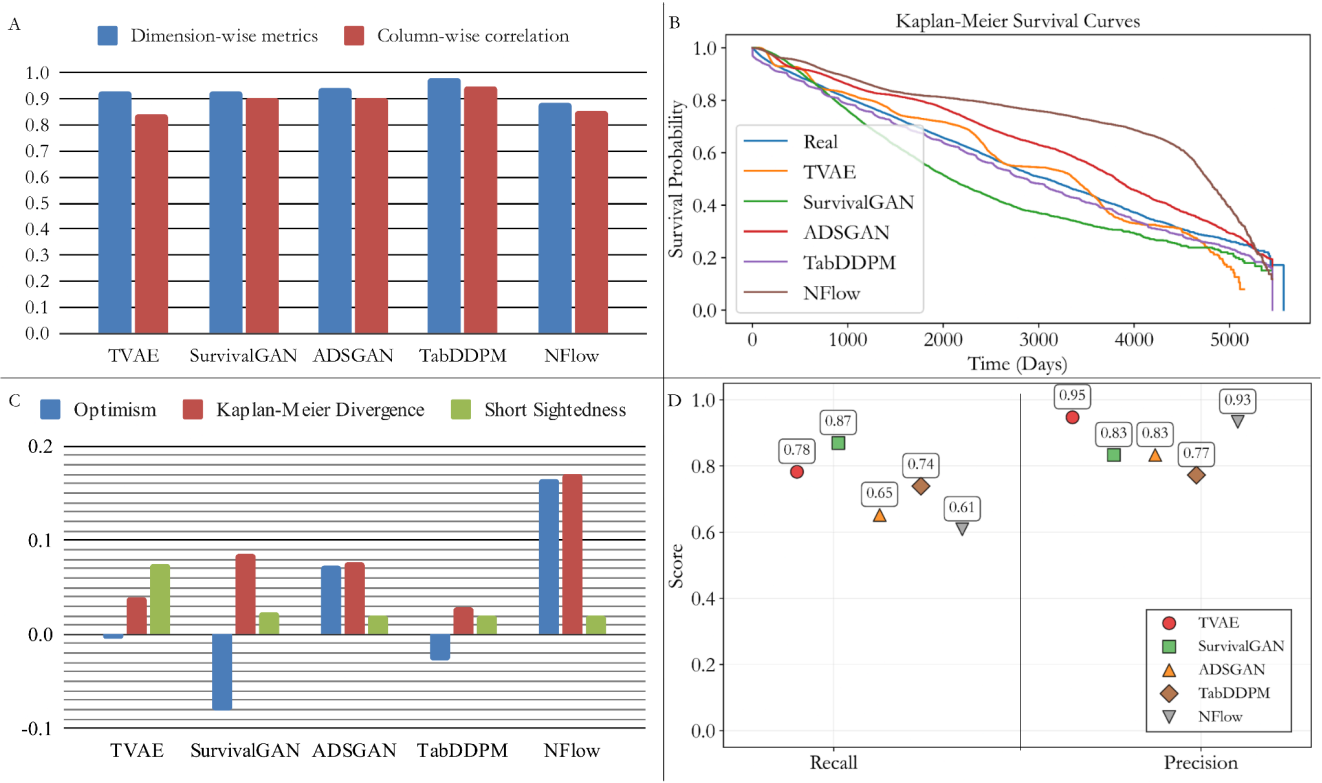

Figure 3: Utility Evaluation of Synthetic Data Generation Models. (A) Assessment of statistical similarity using dimension-wise metrics and column-wise correlations. (B) Kaplan-Meier survival curves comparing synthetic data to the real distribution. (C) Quantitative metrics (Optimism, Kaplan-Meier Divergence, Short Sightedness) measuring survival curve similarity, where values closer to zero are better. (D) Recall and Precision for the preservation of significant prognostic features identified by CoxPH models.

Table 1: Performance metrics (C-index and integrated brier score in parentheses) for survival analysis models trained on datasets with missing values imputed using the MICE method.

|  | CoxPH | DeepHit | DeepSurv | RSF |
|---|---|---|---|---|
| **Real** | 0.74 (0.16) | 0.74 (0.18) | 0.74 (0.16) | 0.73 (0.16) |
| **Train on synthetic test on real (TSTR)** | | | | |
| **ADSGAN** | 0.72 (0.17) | 0.73 (0.18) | 0.73 (0.17) | 0.72 (0.17) |
| **TabDDPM** | 0.7 (0.18) | 0.73 (0.18) | 0.74 (0.16) | 0.73 (0.17) |
| **NFlow** | 0.68 (0.18) | 0.68 (0.2) | 0.68 (0.2) | 0.67 (0.19) |
| **SurvivalGAN** | 0.73 (0.27) | 0.73 (0.2) | 0.71 (0.19) | 0.73 (0.26) |
| **TVAE** | 0.73 (0.19) | 0.73 (0.18) | 0.74 (0.16) | 0.74 (0.19) |
| **Train on real test on synthetic (TRTS)** | | | | |
| **ADSGAN** | 0.69 (0.19) | 0.68 (0.19) | 0.7 (0.18) | 0.69 (0.18) |
| **TabDDPM** | 0.64 (0.18) | 0.68 (0.19) | 0.7 (0.18) | 0.72 (0.17) |
| **NFlow** | 0.59 (0.34) | 0.57 (0.22) | 0.58 (0.22) | 0.58 (0.34) |
| **SurvivalGAN** | 0.72 (0.16) | 0.73 (0.19) | 0.73 (0.18) | 0.72 (0.16) |
| **TVAE** | 0.75 (0.14) | 0.75 (0.17) | 0.75 (0.15) | 0.74 (0.14) |
| **Train on synthetic test on synthetic (TSTS)** | | | | |
| **ADSGAN** | 0.7 (0.16) | 0.7 (0.18) | 0.7 (0.16) | 0.7 (0.16) |
| **TabDDPM** | 0.68 (0.18) | 0.62 (0.19) | 0.75 (0.16) | 0.74 (0.17) |
| **NFlow** | 0.63 (0.15) | 0.64 (0.16) | 0.63 (0.15) | 0.63 (0.15) |
| **SurvivalGAN** | 0.73 (0.17) | 0.75 (0.19) | 0.74 (0.16) | 0.73 (0.17) |
| **TVAE** | 0.75 (0.16) | 0.75 (0.17) | 0.75 (0.15) | 0.75 (0.15) |

The practical utility assessment through TSTR evaluation confirmed these findings (Table 1, Supplementary Tables 4-5). ADSGAN, TabDDPM and TVAE demonstrated survival models to achieve C-indices (0.7-0.75) and Brier scores (0.16-0.19) closely matching those trained on real data (C-index: 0.73-0.76, Brier scores: 0.15-0.18). While SurvivalGAN achieved comparable C-indices (0.71-0.74), it consistently produced inferior Brier scores (0.24-0.27) when evaluated with CoxPH and RSF models. NFlow showed the poorest TSTR performance across multiple model types (C-index: 0.67-0.71, Brier scores: 0.16-0.2). All models show similar performance across imputation methods.

Table 2: Privacy evaluation metrics for synthetic data generation models compared to real data, including AIA Linear, AIA XGB, NNAA, and MIA.

| Method | TVAE | SurvivalGAN | ADSGAN | TabDDPM | NFlow | Real |
|---|---|---|---|---|---|---|
| AIA Linear | 0.512 | 0.503 | 0.502 | 0.506 | 0.471 | 0.515 |
| AIA XGB | 0.516 | 0.5 | 0.499 | 0.518 | 0.472 | 0.527 |
| NNAA | 0.007 | 0 | -0.005 | 0.012 | -0.01 | |
| MIA | 0.5 | 0.5 | 0.5 | 0.501 | 0.5 | |

Privacy evaluation confirmed that all synthetic data generation models (TVAE, SurvivalGAN, ADSGAN, TabDDPM, and NFlow) preserved data confidentiality (Table 2). AIA Linear scores for most models ranged from 0.502-0.512, while AIA XGB scores ranged from 0.499-0.518, both remaining close to the 0.5 baseline and diverging from 0.5 less than real data values (0.515 and 0.527 respectively), indicating that adversaries cannot infer sensitive attributes better than random chance. NFlow demonstrated the weakest privacy protection with scores of 0.471 (AIA Linear) and 0.472 (AIA XGB), furthest from the optimal 0.5 threshold. All models achieved MIA scores of 0.5, confirming no membership inference advantage, while NNAA values near zero (-0.01 to 0.012) indicated minimal overfitting to training data. Additionally, exact match testing revealed zero replicated records across all synthetic datasets, confirming complete absence of direct data leakage.

Table 3: Comparison of public heart failure datasets and our dataset.

| Dataset | Primary Source(s) | Date | Sample Size | Feature Count | Follow-up | Outcome(s) | Population |
|---|---|---|---|---|---|---|---|
| Kaggle Aggregated HF Prediction [38] | Combination of 5 public heart disease datasets | N/A | 918 | 11 | No | Binary heart disease presence (Author claim that majority is HF) | N/A |
| UCI Heart Failure Clinical Records [39,40] | Faisalabad Institute of Cardiology and at the Allied Hospital in Pakistan | Apr–Dec 2015 | 299 | 12 | Yes (median 130 days) | Death, time to death | N/A |
| Zhang et. al. [41,42] | Zigong Fourth People's Hospital in China | 2016-2019 | 2,008 | 168 | Yes (longest 6 month) | Death, readmission, time to death/readmission, time to discharge | IPD patients |
| Johann et. al. [18] (Synthetic) | 5 university hospitals in Germany | 2018-2022 | 1,326 | 11 | Yes | 1 and 3 year Barcelona BioHF mortality scores [43] | OPD and IPD patients |
| Johann et. al. [18] (Synthetic) | 5 university hospitals in Germany | 2018-2022 | 890 | 13 | Yes | 1 and 3 year MAGGIC mortality scores [44] | OPD and IPD patients |
| Our dataset (Synthetic) | Ramathibodi hospital, Mahidol University in Thailand | Jan 2010-Jun 2025 | 12,552 | 35 | Yes (median 42 month, longest 185 month) | Death, time to death | OPD and IPD patients |

Based on the above results, SurvivalGAN demonstrated good statistical similarity and machine learning performance comparable to the real dataset, except for divergent KM curves and elevated Brier scores. We suspect this is caused by dissimilarity in the time-to-death distribution. To address this issue, we conducted an ablation experiment using histogram equalization to match the time-to-death variable ("Days") distribution to that observed in the real training data (Supplementary Figure 8-12). After this adjustment, both SurvivalGAN and TabDDPM produced KM curves that closely matched the real dataset, including subgroup analyses by heart failure subtype (Supplementary Figure 13). Machine learning performance remained consistent with pre-equalization results, while SurvivalGAN's Brier scores improved (Supplementary Table 7). Privacy evaluation confirmed that confidentiality was maintained at the same level (Supplementary Table 8).

## 4. Discussion

This study demonstrates that deep learning-based synthetic data generation can produce high-fidelity, privacy-preserving datasets for heart failure research. Our comprehensive evaluation framework establishes that synthetic data can serve as a reliable substitute for real-world clinical data while maintaining patient confidentiality, contributing to growing evidence that synthetic data represents a viable solution for overcoming traditional privacy-related data sharing limitations.

Our real-world dataset exhibits several clinically relevant characteristics that provide a foundation for synthetic data generation. The CoxPH analysis (Supplementary Figure 14) shows that our real dataset already carries confounding by indication. Several drug variables are significant with a protective sign in the real dataset, including ACEI, ARBs, BB, SGLTi, statins, and higher systolic blood pressure, consistent with established evidence-based therapies for heart failure [45]. Several clinical risks are also significant, for example atrial fibrillation, diabetes, hypertension and higher HbA1c use.

The utility evaluation reveals that synthetic data generated by deep learning models can achieve statistical similarity and predictive performance comparable to real-world clinical data, though with notable variations across generative architectures. Following histogram equalization, SurvivalGAN proved most effective, achieving high fidelity in both statistical similarity and machine learning utility. In contrast, TVAE delivered strong machine learning performance but failed to replicate the original data distribution. The differences in underlying data distributions became more apparent across various training and testing configurations. Models like TabDDPM and ADSGAN produced visually similar data and performed well in the TSTR paradigm. However, their performance dropped in other scenarios, such as TRTS or TSTS. This discrepancy suggests that while some synthetic datasets are useful for specific tasks, they may contain noise or have a different distribution from the real data. Finally, NFlow showed lower performance in both machine learning test and statistical similarity test.

Despite achieving high discrimination (C-index), SurvivalGAN demonstrated poor calibration (Brier score), a discrepancy may be caused by a distributional variation in the synthetic time-to-death variable. This was corrected by applying histogram equalization, a form of domain adaptation, which was previously used in synthetic image data to close the gap between synthetic and real image data [46]. The intervention corrected the distributional shift, improving Brier scores and KM curve similarity without degrading other utility or privacy metrics. This result highlights the efficacy of post-processing for correcting distributional variation in synthetic tabular data, suggesting future work could explore more on domain adaptation for improving synthetic tabular dataset.

The accurate prediction of mortality in patients with HF remains a significant clinical challenge. As highlighted by McMurray and Simpson, many established risk scores have shown diminishing performance in modern patient cohorts, creating a need for new prognostic tools [47]. Addressing this challenge requires robust datasets, and the importance of public data in advancing both heart failure and AI research is well established, with existing datasets having catalyzed significant scientific progress. For example, focus on public HF dataset, the Zhang et al. dataset has supported prediction models [48] and explanatory studies [49], while the Johann et al. synthetic dataset has been instrumental in research on synthetic data quality [50], and the widely used UCI Heart Failure Clinical Records dataset has enabled various predictive modeling studies [51–53]. However, with only five public HF datasets currently available, expanding this resource base could substantially accelerate research progress. Our synthetic dataset addresses this need by offering several distinct advantages, as summarized in Table 3. With 12,552 patients, it significantly exceeds the size of both the real-world dataset (Zhang et al., 2,008 patients) [41] and synthetic dataset (Johann et al., 1,326 patients) [18], enabling more reliable analyses. Furthermore, its extended follow-up duration, with a median survival of 36 months, may support comprehensive studies of long-term disease trajectories and enhanced prognostic modeling capabilities.

While our work has several strengths, we acknowledge that the dataset was generated from a single institution. This origin may embed local demographic characteristics and clinical practices into the data, potentially limiting the generalizability of predictive models trained on it. Furthermore, inherent institutional biases could be encoded in the synthetic records, impacting the fairness of downstream analyses [54]. However, this very limitation underscores a primary goal of our project. By making this large dataset publicly available, we provide the scientific community with a foundational resource. It can be used as a primary dataset for new investigations or as a valuable comparison cohort for researchers seeking to validate the external applicability of their own models, a common practice in modern predictive modeling [55].

Additionally, the 15-year collection period (2010-2025) encompasses major paradigm shifts in heart failure treatment that may cause the synthetic data to reflect heterogeneous and outdated practices, with our additional analysis showing marginally higher performance metrics for more recent data (≥2020) compared to pre-2020 data across both real and synthetic datasets (Supplementary Table 9). The introduction of sacubitril/valsartan post-PARADIGM-HF (2014) [56] and SGLT2i following DAPA-HF (2019) [57] and EMPEROR-Reduced (2020) [58] trials represents paradigm shifts that occurred mid-study. This temporal heterogeneity may create treatment patterns in synthetic data that reflect historical rather than contemporary practice, creating temporal heterogeneity in treatment patterns. Furthermore, The lack of New York Heart Association functional class may influence therapeutic decisions, device eligibility, and prognostic stratification [45]. The high missingness rates for cardiac biomarkers (NT-proBNP: 80%, BNP: 82%) and absence of high-sensitivity troponin further limit comprehensive risk assessment capabilities. These phenotyping gaps may result in synthetic patients lacking the functional and biochemical complexity necessary for realistic clinical decision-making scenarios.

## 5. Conclusion

In conclusion, this study describes the generation and evaluation of a new synthetic dataset for heart failure research. Our findings indicate that it is possible to generate high-quality, privacy-preserving synthetic data that reflects key characteristics of the original dataset. This work contributes to ongoing efforts to address the challenges of data sharing in medical research and aims to provide a useful resource for the scientific community. The code used to implement this research is available at https://github.com/44REAM/Synthetic-Heart-Failure, and the synthetic dataset is publicly available at https://doi.org/10.5281/zenodo.17051668.

# Supplementary materials

**Supplementary A: Additional table and figures.**

**Supplementary** Table1: Summary statistics in real data and synthetic data generated by TVAE, SurvivalGAN, ADSGAN, CTGAN, and DDPM models (continuous variables reported as mean ± standard deviation (missing rate), categorical variables as prevalence). There is no missing data in categorical variables. * indicates no significant difference from the real dataset ($p \geq 0.05$).

|  | Real | TVAE | SurvivalGAN | ADSGAN | TabDDPM | NFlow |
|---|---|---|---|---|---|---|
| HGB | 11.1 ± 2.3 (0.10) | 11.0 ± 1.8 (0.05) | 12.2 ± 2.2 (0.11) | 10.8 ± 2.1 (0.11) | 11.1 ± 2.6* (0.10) | 11.2 ± 2.4* (0.12) |
| Glucose | 139.4 ± 74.5 (0.54) | 124.6 ± 43.7 (0.50) | 112.7 ± 55.9 (0.53) | 108.9 ± 63.1 (0.54) | 153.0 ± 129.8 (0.54) | 141.1 ± 85.3* (0.70) |
| HbA1C (EDTA-blood) | 7.0 ± 1.6 (0.30) | 6.7 ± 1.3 (0.26) | 7.2 ± 1.7 (0.35) | 7.0 ± 1.7 (0.34) | 7.0 ± 1.9 (0.30) | 6.7 ± 1.5 (0.33) |
| Sodium | 135.9 ± 4.8 (0.09) | 136.4 ± 3.7 (0.05) | 135.5 ± 4.4 (0.08) | 134.1 ± 5.3 (0.06) | 135.7 ± 6.3* (0.08) | 134.8 ± 4.7 (0.10) |
| Potassium | 4.6 ± 0.6 (0.08) | 4.6 ± 0.4* (0.03) | 4.8 ± 0.5 (0.08) | 4.6 ± 0.5* (0.07) | 4.7 ± 0.8* (0.07) | 4.5 ± 0.5 (0.09) |
| Blood urea nitrogen | 29.7 ± 22.0 (0.09) | 27.4 ± 17.1* (0.05) | 29.7 ± 20.2 (0.09) | 31.0 ± 23.5 (0.08) | 30.2 ± 26.0* (0.09) | 31.7 ± 22.0 (0.08) |
| Creatinine | 1.8 ± 2.1 (0.05) | 1.5 ± 1.4 (0.01) | 2.1 ± 2.3 (0.05) | 2.2 ± 2.7 (0.05) | 1.8 ± 2.5 (0.05) | 2.1 ± 2.0 (0.05) |
| LDL Cholesterol | 106.5 ± 41.0 (0.19) | 102.0 ± 33.4 (0.10) | 99.1 ± 37.2 (0.21) | 90.3 ± 36.5 (0.20) | 106.9 ± 46.2 (0.18) | 100.0 ± 33.0 (0.21) |
| HR | 89.3 ± 17.5 (0.15) | 89.2 ± 14.3* (0.09) | 86.7 ± 16.0 (0.15) | 86.0 ± 15.3 (0.13) | 90.8 ± 25.7* (0.14) | 92.1 ± 18.5 (0.18) |
| HIGH | 159.7 ± 9.2 (0.29) | 159.4 ± 8.3* (0.23) | 161.9 ± 9.0 (0.28) | 161.5 ± 9.3 (0.32) | 160.0 ± 10.0 (0.30) | 162.1 ± 10.6 (0.27) |
| BW | 65.8 ± 16.3 (0.16) | 64.5 ± 13.3 (0.09) | 65.6 ± 14.4* (0.16) | 67.4 ± 15.2 (0.14) | 65.7 ± 16.7* (0.17) | 71.8 ± 12.7 (0.15) |
| SBP | 148.6 ± 23.9 (0.15) | 147.5 ± 20.4 (0.07) | 155.1 ± 22.9 (0.14) | 148.3 ± 23.4* (0.12) | 149.5 ± 25.4* (0.16) | 141.5 ± 23.2 (0.17) |
| DBP | 81.9 ± 11.2 (0.15) | 82.2 ± 7.9 (0.08) | 84.3 ± 10.1 (0.16) | 84.7 ± 10.3 (0.14) | 82.0 ± 12.5* (0.16) | 82.3 ± 8.9 (0.20) |
| SPO2 | 96.2 ± 3.4 (0.41) | 96.6 ± 2.1 (0.38) | 96.4 ± 2.8 (0.42) | 96.5 ± 3.1 (0.43) | 96.1 ± 5.3* (0.41) | 92.2 ± 13.1 (0.52) |
| NT-proBNP | 5867.3 ± 9521.4 (0.80) | 3864.8 ± 5763.2 (0.81) | 4919.0 ± 7394.8 (0.77) | 8607.4 ± 10793.4 (0.78) | 6247.2 ± 9079.6* (0.82) | 11388.4 ± 12947.0 (0.77) |
| proBNP | 6763.4 ± 9795.4 (0.82) | 5681.9 ± 7650.7 (0.83) | 9460.7 ± 10142.7 (0.79) | 7224.1 ± 8515.6* (0.80) | 7371.6 ± 8976.0* (0.83) | 6142.4 ± 7894.7* (0.88) |
| Age | 67.9 ± 14.2 (0.00) | 69.5 ± 12.3 (0.00) | 68.5 ± 14.3 (0.00) | 67.3 ± 14.6 (0.00) | 66.5 ± 16.3 (0.00) | 62.7 ± 13.8 (0.00) |
| Days | 1578.4 ± 1303.0 (0.00) | 1490.8 ± 1190.7 (0.00) | 1221.3 ± 967.4 (0.00) | 1829.8 ± 1426.7 (0.00) | 1515.1 ± 1270.5 (0.00) | 2337.3 ± 1701.2 (0.00) |
| ACEI | 0.24 | 0.2 | 0.24* | 0.3 | 0.25* | 0.18 |
| ARBs | 0.4 | 0.37 | 0.43 | 0.39* | 0.39* | 0.51 |
| ARNI | 0.05 | 0.03 | 0.07 | 0.06 | 0.05 | 0.22 |
| BB | 0.76 | 0.82 | 0.71 | 0.8 | 0.77* | 0.77 |
| Ivabradine | 0.05 | 0.01 | 0.06 | 0.05* | 0.05* | 0.18 |
| MRA | 0.37 | 0.38 | 0.4 | 0.38* | 0.36* | 0.42 |
| SGLT2i | 0.22 | 0.17 | 0.2 | 0.25 | 0.21* | 0.15 |
| Statin | 0.7 | 0.78 | 0.69* | 0.68 | 0.71 | 0.76 |
| furosemide | 0.54 | 0.52 | 0.59 | 0.56 | 0.55* | 0.65 |
| thiazide | 0.06 | 0.02 | 0.08 | 0.07* | 0.05 | 0.07 |
| HT | 0.61 | 0.68 | 0.62* | 0.65 | 0.62* | 0.78 |
| DM | 0.51 | 0.48 | 0.5 | 0.50* | 0.53 | 0.51* |
| AF | 0.41 | 0.46 | 0.45 | 0.45 | 0.4 | 0.32 |
| CKD | 0.54 | 0.59 | 0.53* | 0.5 | 0.55* | 0.68 |
| dead | 0.36 | 0.31 | 0.36* | 0.32 | 0.38 | 0.29 |
| type | HFpEF: 0.7 HFmrEF: 0.17 HFrEF: 0.12 | HFpEF: 0.78 HFmrEF: 0.15 HFrEF: 0.07 | HFpEF: 0.66 HFmrEF: 0.19 HFrEF: 0.16 | HFpEF: 0.69 HFmrEF: 0.19 HFrEF: 0.12 | HFpEF: 0.72 HFmrEF: 0.16 HFrEF: 0.12* | HFpEF: 0.63 HFmrEF: 0.27 HFrEF: 0.1 |
| Gender | 0.5 | 0.48 | 0.50* | 0.51* | 0.50* | 0.48 |

**Supplementary Table 2:** Clinically plausible ranges.

| variable | min | max |
|---|---|---|
| HGB | 3 | 20 |
| Glucose | 10 | 1000 |
| HbA1C (EDTA-blood) | 4 | 15 |
| Sodium | 60 | 160 |
| Potassium | 1 | 10 |
| Blood urea nitrogen | 5 | 200 |
| Creatinine | 0.1 | 30 |
| LDL Cholesterol | 20 | 500 |
| HR | 20 | 300 |
| HIGH | 100 | 250 |
| BW | 20 | 200 |
| SBP | 40 | 250 |
| DBP | 20 | 200 |
| SP02 | 20 | 100 |
| NT-proBNP | 1 | 50000 |
| proBNP | 1 | 50000 |

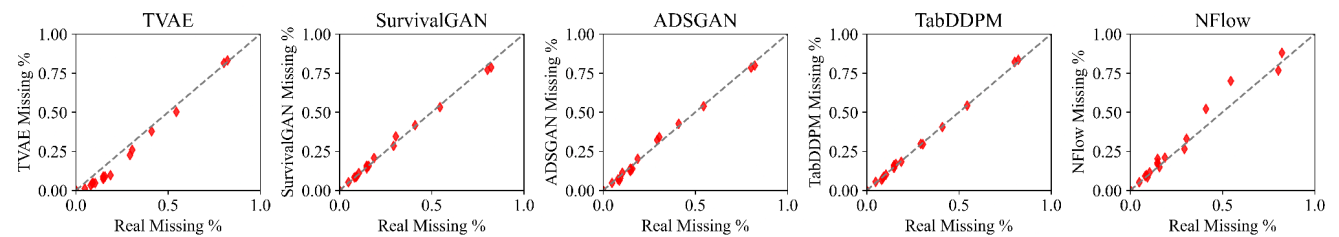

**Supplementary Figure 1:** Scatter plots comparing the proportion of missing values of continuous features between the real dataset (x-axis) and the synthetic datasets generated by five different models (y-axis). The dashed line indicates a perfect correspondence.

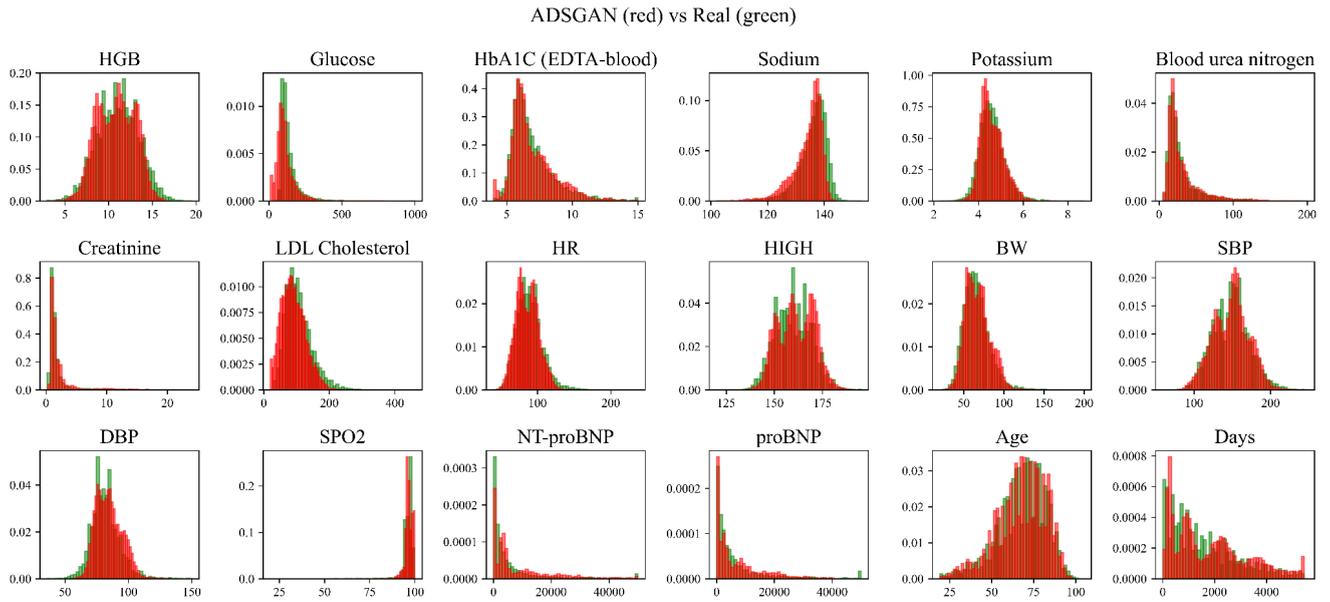

**Supplementary Figure 2:** Comparison of data distributions for continuous variables between the real dataset (green) and the synthetic data generated by a ADSGAN (red).

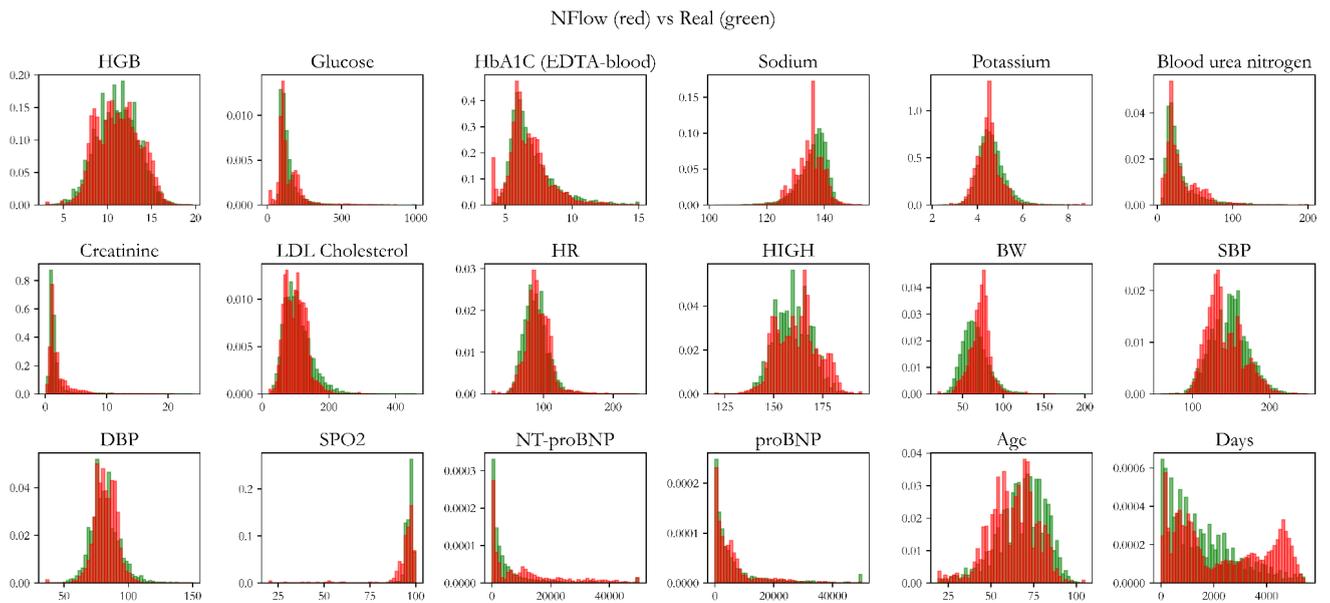

**Supplementary Figure 3:** Comparison of data distributions for continuous variables between the real dataset (green) and the synthetic data generated by a NFlow (red).

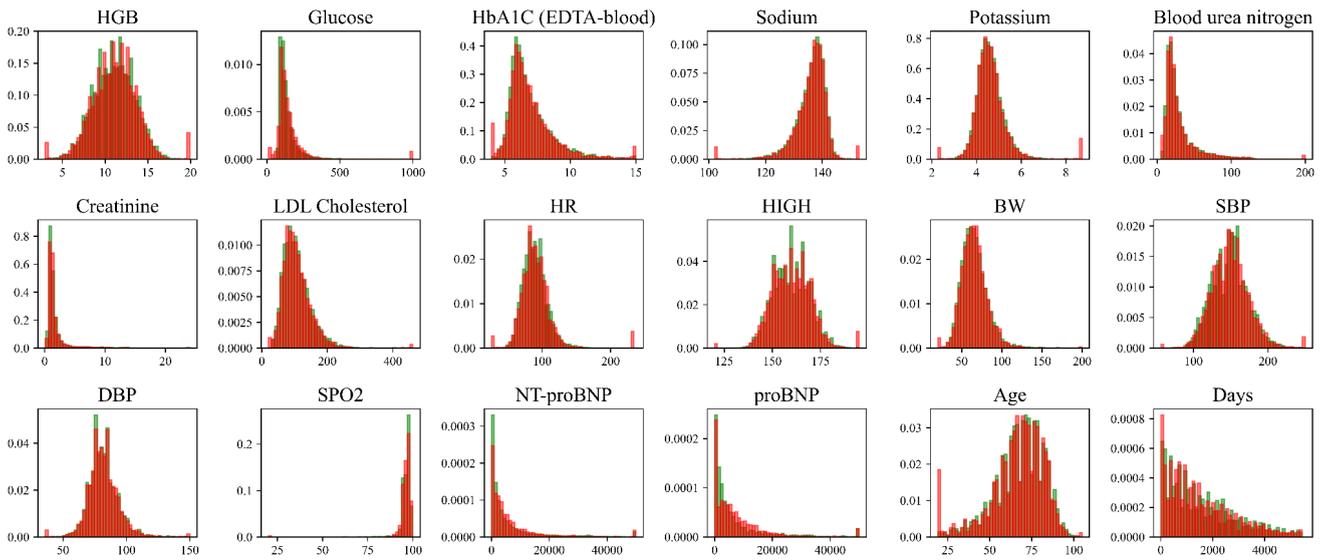

**Supplementary Figure 4:** Comparison of data distributions for continuous variables between the real dataset (green) and the synthetic data generated by a TabDDPM (red).

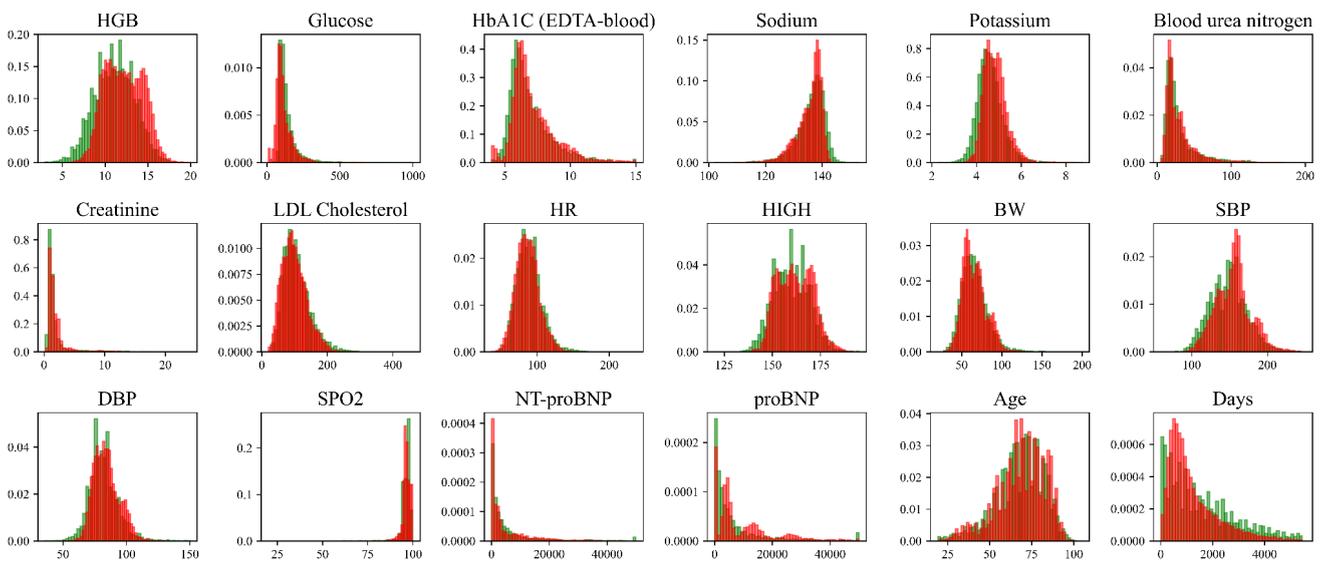

**Supplementary Figure 5:** Comparison of data distributions for continuous variables between the real dataset (green) and the synthetic data generated by a SurvivalGAN (red).

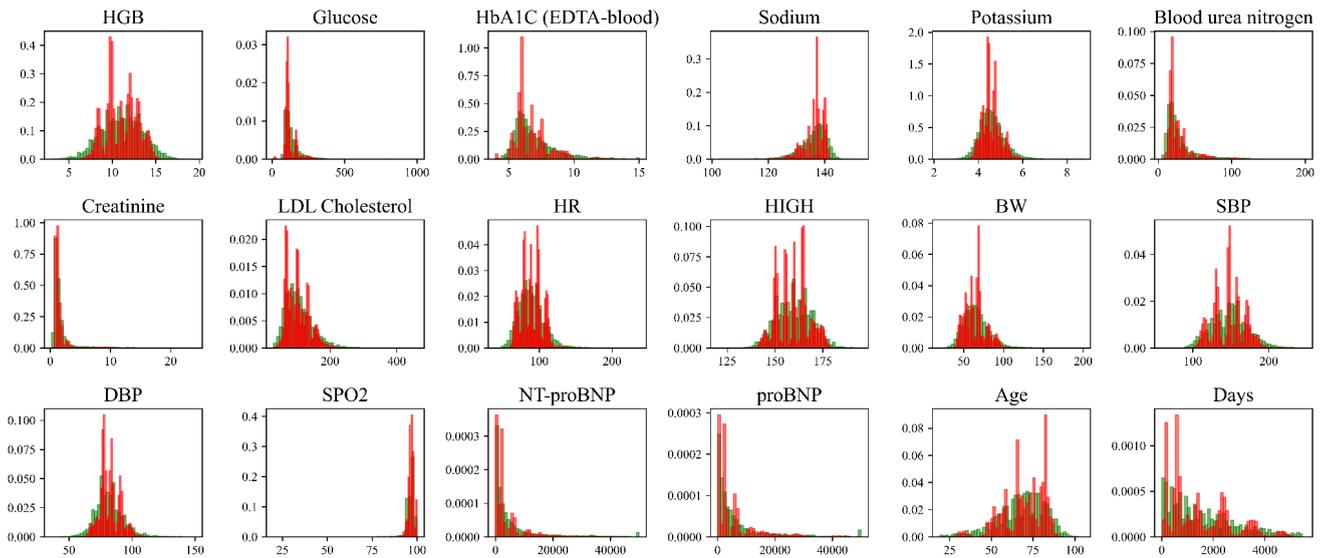

**Supplementary Figure 6:** Comparison of data distributions for continuous variables between the real dataset (green) and the synthetic data generated by a TVAE (red).

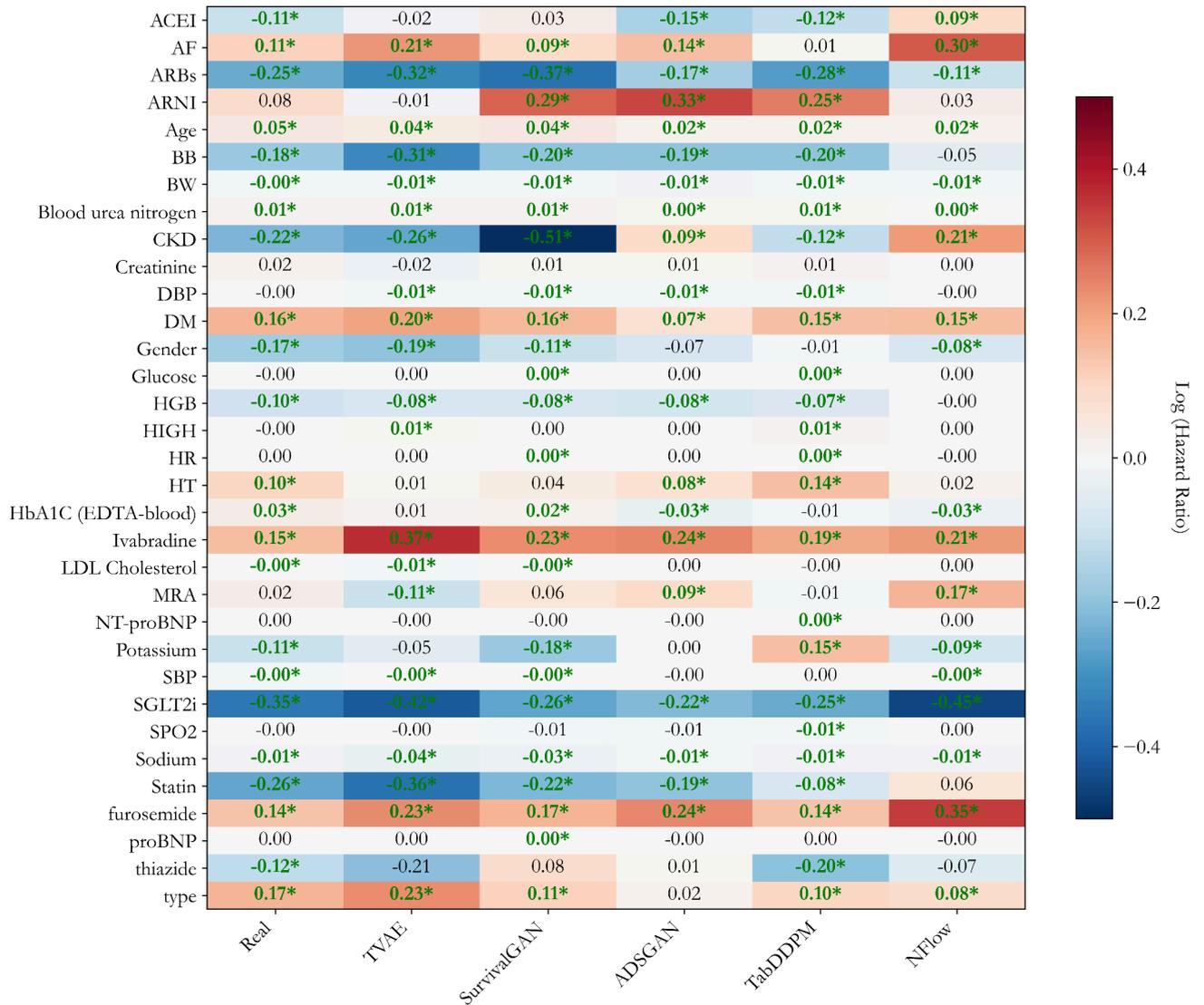

**Supplementary Figure 7:** Cox coefficient heatmap.

**Supplementary Table 3**: Hyperparameter settings for survival models.

| Model | Hyperparameters | Value/Setting |
|---|---|---|
| Cox proportional hazard (CoxPH) | L1 ratio | [0,0.5,1] |
| | Penalty | [ 0.1, 1] |
| Random survival forest (RSF) | N estimators | [5, 20, 50] |
| | Max depth | [2, 5, 10] |
| | Min samples split | [2, 5, 10] |
| | Min samples leaf | [1, 2, 4] |
| Deep learning (DeepSurv and DeepHit) | Batch size | 256 |
| | Max epochs | 1000 |
| | Early stopping patience | 50 |
| | Learning rate | 0.0001 |
| | Time discretization bins (only DeepHit) | 10 |

**Supplementary Table 4**: Performance metrics (C-index and Brier score in parentheses) for survival analysis models trained on datasets with missing values imputed using the median method.

| | CoxPH | DeepHit | DeepSurv | RSF |
|---|---|---|---|---|
| **Real** | 0.74 (0.16) | 0.73 (0.17) | 0.76 (0.15) | 0.75 (0.15) |
| **Train on synthetic test on real (TSTR)** | | | | |
| **ADSGAN** | 0.73 (0.16) | 0.73 (0.18) | 0.74 (0.16) | 0.73 (0.16) |
| **TabDDPM** | 0.72 (0.18) | 0.74 (0.18) | 0.75 (0.16) | 0.74 (0.16) |
| **NFlow** | 0.68 (0.18) | 0.69 (0.2) | 0.68 (0.2) | 0.71 (0.18) |
| **SurvivalGAN** | 0.74 (0.27) | 0.74 (0.2) | 0.72 (0.19) | 0.74 (0.24) |
| **TVAE** | 0.74 (0.18) | 0.73 (0.18) | 0.75 (0.16) | 0.74 (0.18) |
| **Train on real test on synthetic (TRTS)** | | | | |
| **ADSGAN** | 0.7 (0.19) | 0.68 (0.19) | 0.69 (0.2) | 0.71 (0.19) |
| **TabDDPM** | 0.65 (0.17) | 0.7 (0.19) | 0.71 (0.17) | 0.72 (0.16) |
| **NFlow** | 0.58 (0.38) | 0.57 (0.22) | 0.58 (0.29) | 0.58 (0.4) |
| **SurvivalGAN** | 0.73 (0.16) | 0.74 (0.19) | 0.74 (0.17) | 0.73 (0.15) |
| **TVAE** | 0.75 (0.14) | 0.75 (0.17) | 0.75 (0.15) | 0.74 (0.14) |
| **Train on synthetic test on synthetic (TSTS)** | | | | |
| **ADSGAN** | 0.71 (0.15) | 0.71 (0.17) | 0.71 (0.15) | 0.72 (0.15) |
| **TabDDPM** | 0.69 (0.18) | 0.64 (0.19) | 0.76 (0.16) | 0.75 (0.16) |
| **NFlow** | 0.63 (0.15) | 0.64 (0.15) | 0.63 (0.15) | 0.64 (0.15) |
| **SurvivalGAN** | 0.74 (0.17) | 0.77 (0.18) | 0.76 (0.15) | 0.75 (0.16) |
| **TVAE** | 0.75 (0.16) | 0.76 (0.16) | 0.76 (0.15) | 0.76 (0.15) |

**Supplementary Table 5**: Performance metrics (C-index and Brier score in parentheses) for survival analysis models trained on datasets with missing values imputed using the Hyperimpute method.

| | CoxPH | DeepHit | DeepSurv | RSF |
|---|---|---|---|---|
| **Real** | 0.74 (0.16) | 0.73 (0.18) | 0.76 (0.15) | 0.75 (0.15) |
| **Train on synthetic test on real (TSTR)** | | | | |
| **ADSGAN** | 0.73 (0.16) | 0.74 (0.18) | 0.74 (0.16) | 0.73 (0.16) |
| **TabDDPM** | 0.72 (0.18) | 0.74 (0.18) | 0.75 (0.16) | 0.74 (0.16) |
| **NFlow** | 0.68 (0.18) | 0.69 (0.2) | 0.68 (0.2) | 0.71 (0.18) |
| **SurvivalGAN** | 0.74 (0.27) | 0.74 (0.2) | 0.72 (0.19) | 0.74 (0.24) |
| **TVAE** | 0.74 (0.18) | 0.73 (0.18) | 0.75 (0.16) | 0.74 (0.18) |
| **Train on real test on synthetic (TRTS)** | | | | |
| **ADSGAN** | 0.7 (0.19) | 0.67 (0.19) | 0.68 (0.22) | 0.7 (0.2) |
| TabDDPM | 0.65 (0.17) | 0.7 (0.19) | 0.71 (0.17) | 0.72 (0.16) |
| NFlow | 0.58 (0.38) | 0.57 (0.22) | 0.58 (0.29) | 0.58 (0.4) |
| SurvivalGAN | 0.73 (0.16) | 0.74 (0.19) | 0.74 (0.17) | 0.73 (0.15) |
| TVAE | 0.75 (0.14) | 0.75 (0.17) | 0.75 (0.15) | 0.74 (0.14) |
| **Train on synthetic test on synthetic (TSTS)** | | | | |
| ADSGAN | 0.7 (0.15) | 0.71 (0.17) | 0.71 (0.15) | 0.7 (0.15) |
| TabDDPM | 0.69 (0.18) | 0.64 (0.19) | 0.76 (0.16) | 0.75 (0.16) |
| NFlow | 0.63 (0.15) | 0.64 (0.15) | 0.63 (0.15) | 0.64 (0.15) |
| SurvivalGAN | 0.74 (0.17) | 0.77 (0.18) | 0.76 (0.15) | 0.75 (0.16) |
| TVAE | 0.75 (0.16) | 0.76 (0.16) | 0.76 (0.15) | 0.76 (0.15) |

**Supplementary Table 6:** Variables definition. For binary variables, 0 = absent, 1 = present

| Variable | Full Name | Type | Unit | Category |
|---|---|---|---|---|
| HGB | Hemoglobin | Continuous | g/dL | Laboratory Values |
| Glucose | Blood Glucose | Continuous | mg/dL | Laboratory Values |
| HbA1C (EDTA-blood) | HbA1C | Continuous | % | Laboratory Values |
| Sodium | Serum Sodium | Continuous | mEq/L | Laboratory Values |
| Potassium | Serum Potassium | Continuous | mEq/L | Laboratory Values |
| Blood urea nitrogen | Blood Urea Nitrogen (BUN) | Continuous | mg/dL | Laboratory Values |
| Creatinine | Serum Creatinine | Continuous | mg/dL | Laboratory Values |
| LDL Cholesterol | Low-Density Lipoprotein | Continuous | mg/dL | Laboratory Values |
| HR | Heart Rate | Continuous | bpm | Vital Signs |
| HIGH | Height | Continuous | cm | Vital Signs |
| BW | Body Weight | Continuous | kg | Vital Signs |
| SBP | Systolic Blood Pressure | Continuous | mmHg | Vital Signs |
| DBP | Diastolic Blood Pressure | Continuous | mmHg | Vital Signs |
| SPO2 | Oxygen Saturation | Continuous | % | Vital Signs |
| NT-proBNP | N-terminal pro B-type Natriuretic Peptide | Continuous | pg/mL | Laboratory Values |
| proBNP | Pro B-type Natriuretic Peptide | Continuous | pg/mL | Laboratory Values |
| Age | Age | Continuous | years | Demographics |
| Gender | Gender | Binary | 0/1 | Demographics |
| ACEI | ACE Inhibitors | Binary | 0/1 | Medications |
| ARBs | Angiotensin Receptor Blockers | Binary | 0/1 | Medications |
| ARNI | Angiotensin Receptor-Neprilysin Inhibitor | Binary | 0/1 | Medications |
| BB | Beta Blockers | Binary | 0/1 | Medications |
| Ivabradine | Ivabradine | Binary | 0/1 | Medications |
| MRA | Mineralocorticoid Receptor Antagonists | Binary | 0/1 | Medications |
| SGLT2i | SGLT2 Inhibitors | Binary | 0/1 | Medications |
| Statin | Statin | Binary | 0/1 | Medications |

| furosemide | Furosemide | Binary | 0/1 | Medications |
| thiazide | Thiazide Diuretics | Binary | 0/1 | Medications |
| HT | Hypertension | Binary | 0/1 | Co-morbidity |
| DM | Diabetes Mellitus | Binary | 0/1 | Co-morbidity |
| AF | Atrial Fibrillation | Binary | 0/1 | Co-morbidity |
| CKD | Chronic Kidney Disease | Binary | 0/1 | Co-morbidity |
| dead | Death Status | Binary | 0/1 | Outcomes |
| Days | Time to Death/Censoring | Continuous | days | Outcomes |
| type | Heart Failure type | Categorical | 0=HFpEF 1=HFmrEF 2=HFrEF | Co-morbidity |

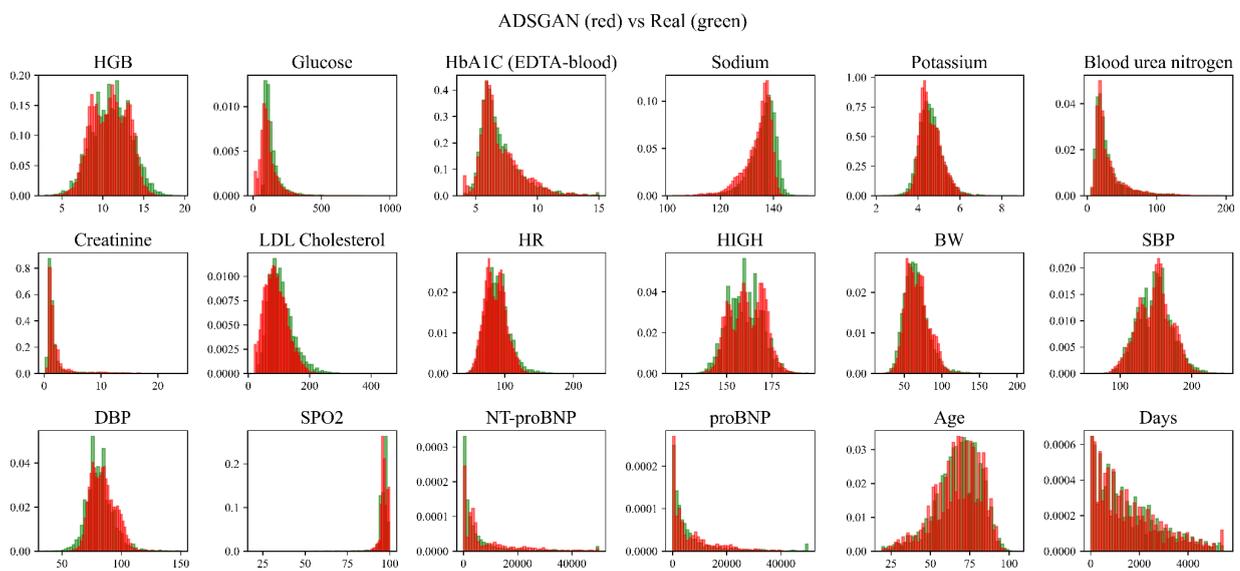

**Supplementary Figure 8:** Comparison of data distributions for continuous variables between the real dataset (green) and the synthetic data generated by a ADSGAN (red). The "Days" variable was histogram equalized to match the real dataset distribution.

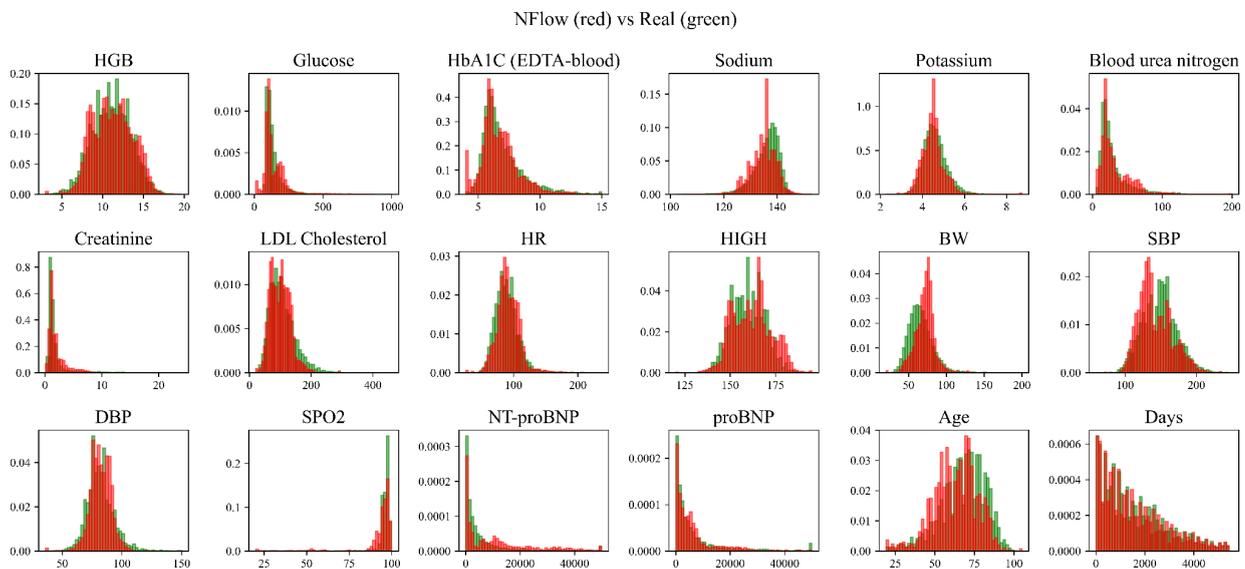

**Supplementary Figure 9:** Comparison of data distributions for continuous variables between the real dataset (green) and the synthetic data generated by a NFlow (red). The "Days" variable (time-to-death) was histogram equalized to match the real dataset distribution.

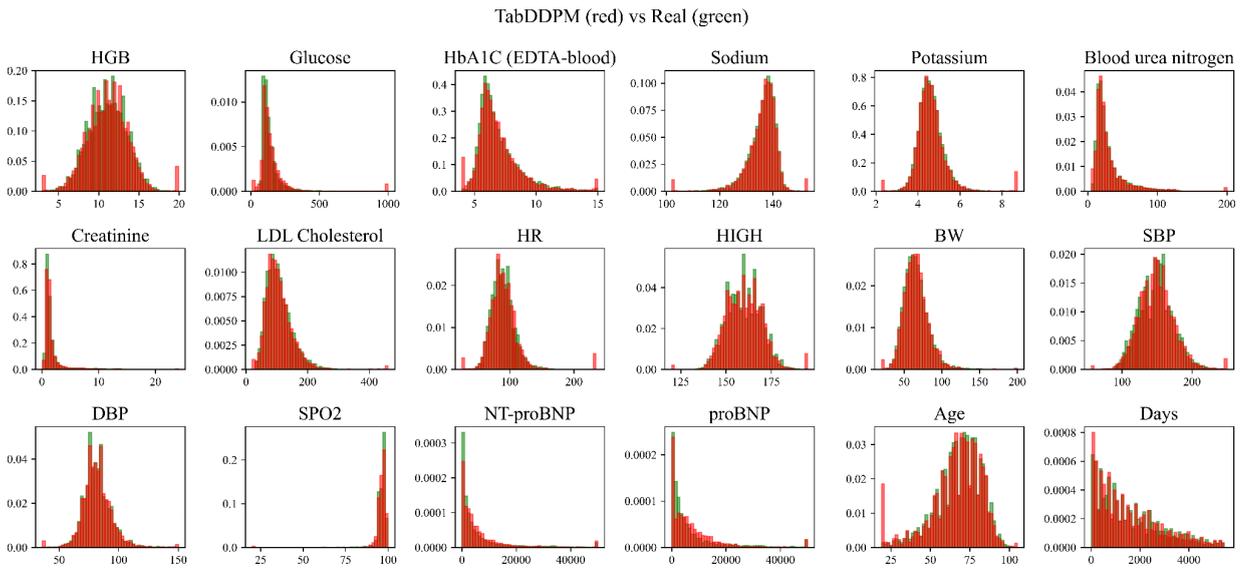

**Supplementary Figure 10:** Comparison of data distributions for continuous variables between the real dataset (green) and the synthetic data generated by a TabDDPM (red). The "Days" variable (time-to-death) was histogram equalized to match the real dataset distribution.

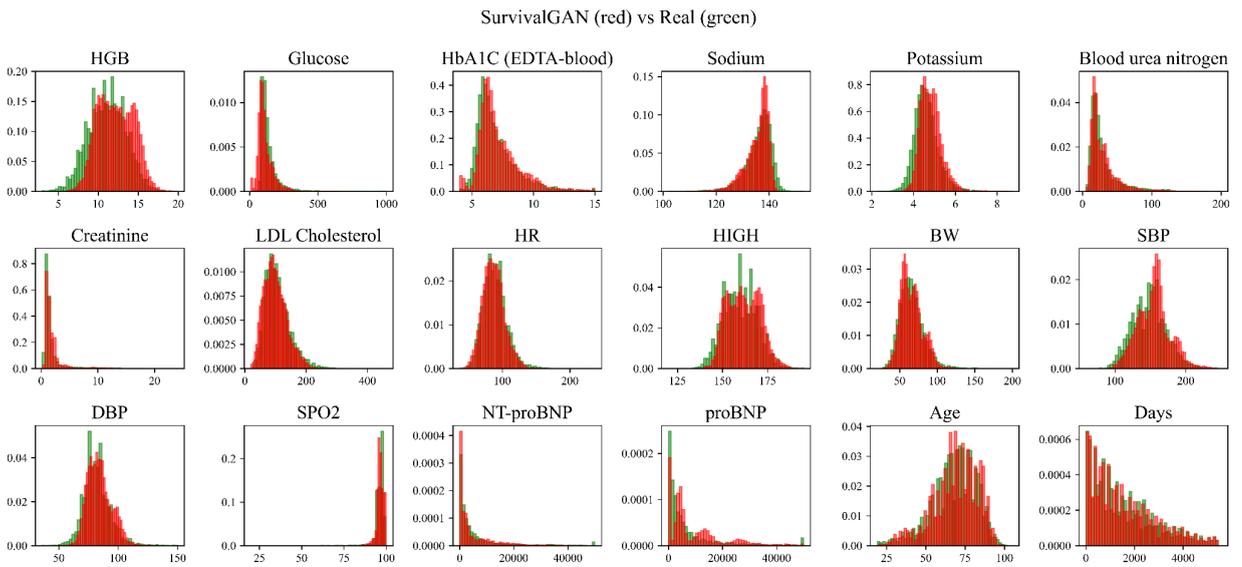

**Supplementary Figure 11:** Comparison of data distributions for continuous variables between the real dataset (green) and the synthetic data generated by a SurvivalGAN (red). The "Days" variable (time-to-death) was histogram equalized to match the real dataset distribution.

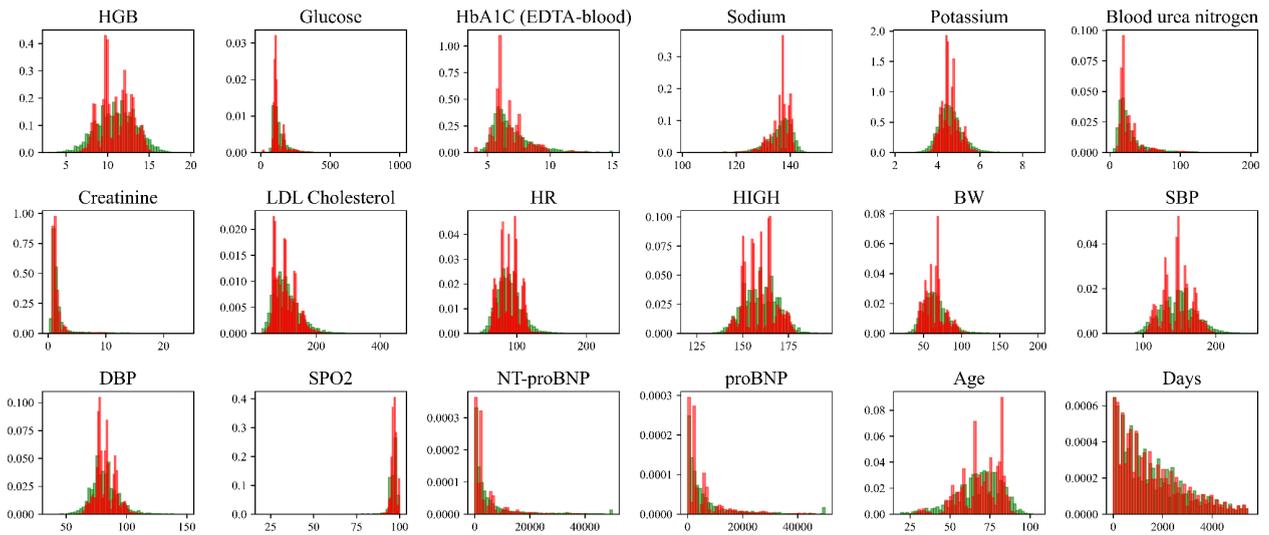

**Supplementary Figure 12:** Comparison of data distributions for continuous variables between the real dataset (green) and the synthetic data generated by a TVAE (red). The "Days" variable (time-to-death) was histogram equalized to match the real dataset distribution.

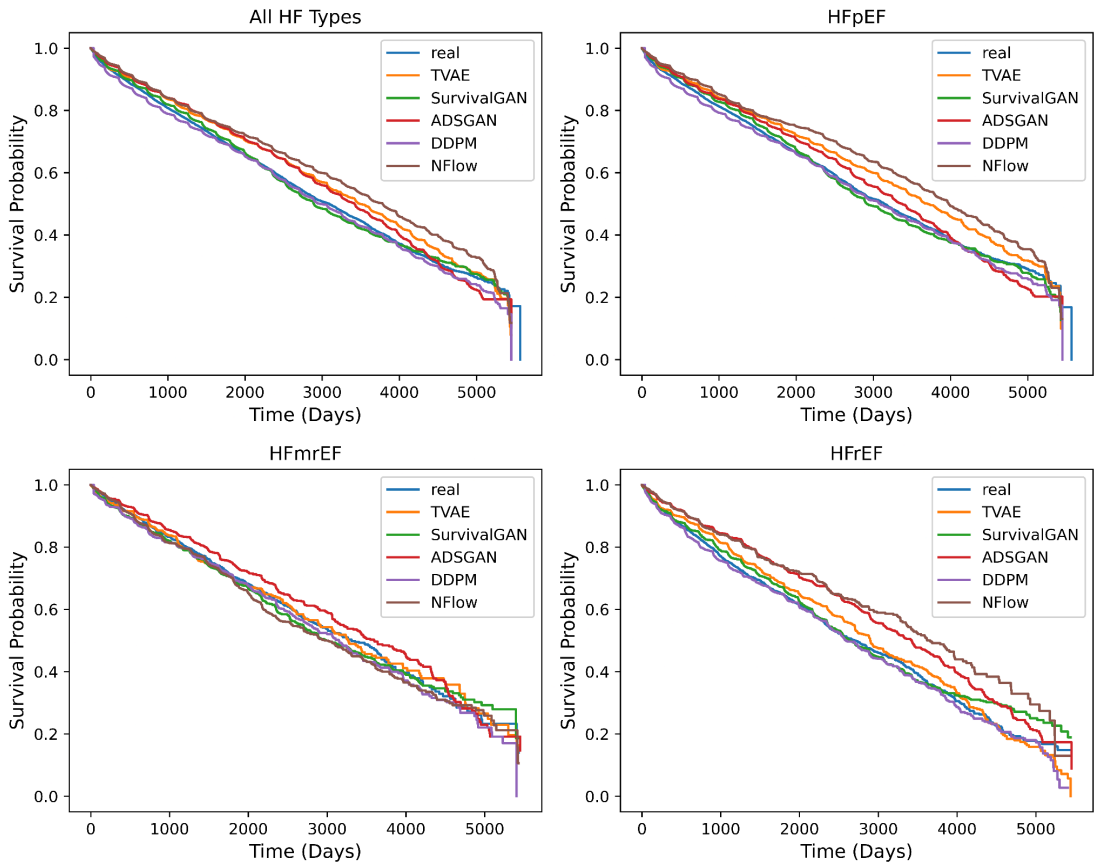

**Supplementary Figure 13:** Kaplan-Meier survival curves comparing synthetic data generation methods against real data across different heart failure subtypes. The "Days" variable (time-to-death) was histogram equalized to match the real dataset distribution before fitted to the KM curve.

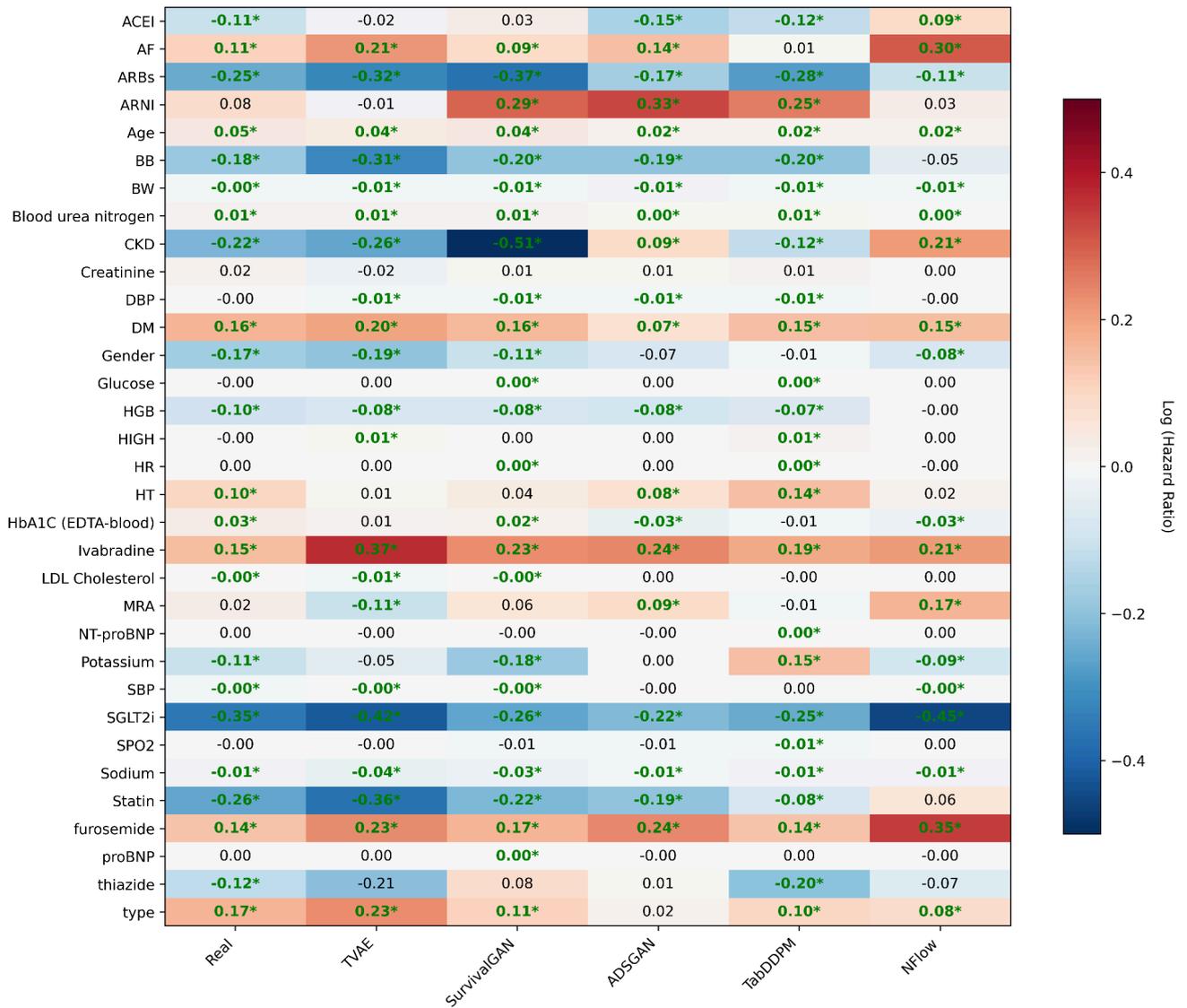

**Supplementary Figure 14:** Cox coefficient heatmap. The "Days" variable (time-to-death) was histogram equalized to match the real dataset distribution before fitted to the CoxPH model.

**Supplementary Table 7**: Performance metrics (C-index and Brier score in parentheses) for survival analysis models trained on datasets with missing values imputed using the MICE method. The "Days" variable (time-to-death) was histogram equalized to match the real dataset distribution before developing the survival model.

|  | **CoxPH** | **DeepHit** | **DeepSurv** | **RSF** |
|---|---|---|---|---|
| **Real** | 0.74 (0.16) | 0.74 (0.18) | 0.74 (0.16) | 0.73 (0.16) |
| **Train on synthetic test on real (TSTR)** | | | | |
| **ADSGAN** | 0.72 (0.18) | 0.72 (0.18) | 0.72 (0.17) | 0.72 (0.18) |
| **TabDDPM** | 0.7 (0.18) | 0.73 (0.18) | 0.74 (0.16) | 0.73 (0.17) |
| **NFlow** | 0.68 (0.19) | 0.68 (0.19) | 0.68 (0.18) | 0.67 (0.2) |
| **SurvivalGAN** | 0.73 (0.17) | 0.72 (0.18) | 0.74 (0.17) | 0.73 (0.17) |
| **TVAE** | 0.73 (0.18) | 0.73 (0.18) | 0.74 (0.17) | 0.74 (0.18) |
| **Train on real test on synthetic (TRTS)** | | | | |
| **ADSGAN** | 0.69 (0.16) | 0.68 (0.18) | 0.69 (0.17) | 0.69 (0.16) |
| **TabDDPM** | 0.64 (0.19) | 0.69 (0.19) | 0.69 (0.18) | 0.72 (0.17) |
| **NFlow** | 0.59 (0.17) | 0.57 (0.18) | 0.58 (0.2) | 0.58 (0.17) |
| **SurvivalGAN** | 0.72 (0.16) | 0.74 (0.18) | 0.72 (0.16) | 0.72 (0.16) |
| TVAE | 0.75 (0.15) | 0.75 (0.17) | 0.75 (0.15) | 0.74 (0.15) |
| **Train on synthetic test on synthetic (TSTS)** | | | | |
| ADSGAN | 0.7 (0.16) | 0.7 (0.18) | 0.69 (0.17) | 0.7 (0.16) |
| TabDDPM | 0.68 (0.19) | 0.63 (0.19) | 0.75 (0.17) | 0.74 (0.17) |
| NFlow | 0.63 (0.16) | 0.63 (0.17) | 0.63 (0.16) | 0.63 (0.16) |
| SurvivalGAN | 0.73 (0.16) | 0.74 (0.18) | 0.74 (0.16) | 0.73 (0.16) |
| TVAE | 0.75 (0.16) | 0.76 (0.17) | 0.75 (0.15) | 0.75 (0.15) |

**Supplementary Table 8:** Privacy evaluation metrics for synthetic data generation models compared to real data, including AIA Linear, AIA XGB, NNAA, and MIA. The "Days" variable (time-to-death) was histogram equalized to match the real dataset distribution before privacy testing.

| Method | TVAE | SurvivalGAN | ADSGAN | TabDDPM | NFlow | Real |
|---|---|---|---|---|---|---|
| AIA Linear | 0.512 | 0.504 | 0.501 | 0.506 | 0.471 | 0.515 |
| AIA XGB | 0.516 | 0.502 | 0.499 | 0.519 | 0.471 | 0.527 |
| NNAA | 0.008 | -0.002 | -0.002 | 0.012 | -0.008 | |
| MIA | 0.5 | 0.5 | 0.5 | 0.501 | 0.5 | |

**Supplementary Table 9:** Performance metrics (C-index and Brier score in parentheses) for survival analysis models trained on datasets with missing values imputed using the MICE method. The "Days" variable (time-to-death) was histogram equalized to match the real dataset distribution before developing the survival model. Results are stratified by temporal cohorts: recent data from 2020 onwards (n=1,081 patients) and historical data pre-2020 (n=1,430 patients).

| | | CoxPH | DeepHit | DeepSurv | RSF |
|---|---|---|---|---|---|
| **Train on real test on real (TRTR)** | | | | | |
| Real | ≥2020 | 0.76 (0.1) | 0.76 (0.14) | 0.76 (0.13) | 0.74 (0.1) |
| | <2020 | 0.72 (0.26) | 0.73 (0.19) | 0.73 (0.17) | 0.72 (0.25) |
| **Train on synthetic test on real (TSTR)** | | | | | |
| ADSGAN | ≥2020 | 0.75 (0.11) | 0.75 (0.14) | 0.74 (0.14) | 0.74 (0.11) |
| | <2020 | 0.7 (0.28) | 0.71 (0.19) | 0.7 (0.17) | 0.7 (0.28) |
| TabDDPM | ≥2020 | 0.73 (0.11) | 0.75 (0.14) | 0.76 (0.13) | 0.74 (0.1) |
| | <2020 | 0.68 (0.27) | 0.73 (0.19) | 0.72 (0.17) | 0.71 (0.26) |
| NFlow | ≥2020 | 0.72 (0.12) | 0.71 (0.15) | 0.69 (0.15) | 0.7 (0.12) |
| | <2020 | 0.67 (0.29) | 0.66 (0.2) | 0.66 (0.18) | 0.66 (0.3) |
| SurvivalGAN | ≥2020 | 0.75 (0.1) | 0.7 (0.14) | 0.75 (0.13) | 0.75 (0.1) |
| | <2020 | 0.71 (0.27) | 0.7 (0.19) | 0.71 (0.18) | 0.7 (0.27) |
| TVAE | ≥2020 | 0.76 (0.1) | 0.75 (0.14) | 0.76 (0.13) | 0.75 (0.1) |
| | <2020 | 0.71 (0.28) | 0.71 (0.19) | 0.72 (0.18) | 0.72 (0.28) |

**Supplementary B: Synthetic Data Generation**

To account for missing values in the real dataset, we began by creating a binary indicator column for each feature containing nulls to flag the location of missing data (1 = missing, 0 = present). We then imputed the training data using the MICE method. The MICE model was trained exclusively on the feature set, excluding outcome variables like death status and time to death. Subsequent preprocessing steps, such as data normalization, were performed using the default settings within the Synthcity library.

After data generation, we applied a filtering step to ensure the clinical and logical validity of the synthetic records. This process automatically removed any data that violated predefined constraints, such as a systolic blood pressure (SBP) value being lower than the diastolic blood pressure (DBP) or values falling outside plausible physiological ranges (see Supplementary Table 2). The original missingness patterns were then reintroduced into the complete synthetic dataset. Using the indicator columns created earlier, we replaced imputed values with nulls where the original data was missing. Finally, the indicator columns were removed.

**Supplementary C: Machine learning for survival analysis**

For all datasets, we employed a stratified split to ensure balanced representation across survival outcomes, with data partitioned into training (70%), validation (10%), and test (20%) sets. The combined training and validation sets (80%) corresponded exactly to the training data used for synthetic data generation, ensuring no data leakage between real and synthetic data evaluation phases.

Prior to model training, missing values were handled within each data source. For the real dataset, an imputation model was fitted on the combined training and validation sets of the real dataset and then used to impute missing values in the test set of the real dataset. Likewise, for each synthetic dataset, a separate imputation model was fitted on its training and validation sets of synthetic dataset and applied only to its corresponding test set of synthetic dataset. Continuous features were standardized using z-score normalization (mean = 0, standard deviation = 1). The normalization statistics (mean and standard deviation) were always calculated from the training data of the specific experiment and then applied to the test set being evaluated. Categorical features were already binary encoded (0 and 1) and used directly without further preprocessing.

We employed distinct optimization strategies based on model complexity. Traditional machine learning models (Cox Proportional Hazards and Random Survival Forest) underwent systematic hyperparameter tuning using grid search on the validation set, while deep learning models (DeepSurv and DeepHit) used established fixed hyperparameters.

Deep learning models utilized a feed forward network architecture. DeepHit uses a hidden layer with layer dimensions of [3N, 5N, 3N], where N represents the number of input features. All networks were trained using the Adam optimizer with early stopping based on validation loss monitoring to prevent overfitting. Training terminated automatically when validation loss showed no improvement for 20 consecutive epochs, with final model selection based on the checkpoint achieving the lowest validation loss during training. Models were implemented using established Python libraries: Cox Proportional Hazards with lifelines, Random Survival Forest with scikit-survival, and both DeepSurv and DeepHit with PyCox. Detailed hyperparameters are provided in **Supplementary Table 3**.

**Supplementary D: Dataset**

Our study uses a dataset of a real-world retrospective cohort of 12,552 patients with HF, collected at Ramathibodi Hospital between January 2010 and June 2025. The study population was identified through hospital records and electronic health databases, including ICD-10 codes, medication records, and echocardiographic reports. Supplementary information such as medication prescriptions, comorbidities, laboratory results, and procedures was extracted from the hospital information system. Ethical approval for this study was obtained from the Institutional Review Board of the Faculty of Medicine, Ramathibodi Hospital prior to initiation of the research (MURA2025/121).

The identification of patients with heart failure was rigorously validated using multiple criteria. This created initial pools of patients identified via ICD-10 codes (n=16,472 for I50, I10, I11.0, I13.0, I13.2), prescription records for key heart failure medications (n=201,230), or the presence of an echocardiogram report (n=59,009).

From these initial pools, a cohort was constructed by identifying patients who met specific combinations of these criteria, ranging from a single medication to a full therapeutic regimen. A key high-specificity criterion was the four-drug Guideline-Directed Medical Therapy (GDMT) group, which is used to identify patients with a high probability of having heart failure. This group consists of patients concurrently prescribed all four medication classes: a renin-angiotensin system inhibitor (e.g., ACEI, ARB, or ARNI), a beta-blocker, a mineralocorticoid receptor antagonist (MRA), and an SGLT2 inhibitor. This specific medication profile, along with less stringent ones like being on 2-3 main medication groups, was used to select patients from larger groups that combined ICD-10 codes with medication records (n=10,901) or matched medication records to at least 3 groups of GDMT or any of specific medication(ARNI, vericiguat) (n=7,938). Patients identified solely through ICD-10 codes (n=5,571) were also included for further validation.

A crucial validation step was then applied to all these preliminary groups: each patient's record was checked for an available echocardiogram report database. The left ventricular ejection fraction was extracted from cardiologist free text report by regex method or Bi-plane LVEF parameter or Teichholz LVEF parameter or manual chart review. which left ventricular ejection fraction values were extracted to classify patients into categories of preserved (n=9,030), reduced (n=2,147), or mildly reduced (n=1,557) ejection fraction. Patients lacking this report were excluded. This systematic filtering and validation process yielded the final cohort of 12,734 heart failure subjects.

To minimize the rate of missing data, we employed a forward-fill imputation technique, tailoring the time window to each variable's clinical nature. Medication statuses were carried forward for up to 120 days, while more stable parameters like height, weight, and ejection fraction were imputed over a 365-day period. A 180-day window was applied to all remaining variables. Following imputation, the dataset was temporally aggregated into six-month intervals to normalize the time points for analysis. The exact dates of primary outcome events (death) were explicitly preserved to ensure analytical accuracy. Finally, patients who died within 180 days were excluded due to the imputation strategy, yielding a final analytic cohort of 12,552 patients.